\documentclass[table]{article} 
\usepackage{iclr2016_conference,times}
\usepackage{hyperref}
\usepackage{url}
\usepackage{epsfig}
\usepackage{graphicx}
\usepackage{amsmath}
\usepackage{amssymb}
\usepackage{algorithmic}
\usepackage{mathrsfs}

\usepackage{booktabs}
\usepackage{xcolor}
\usepackage{multirow}
\usepackage{pbox, calc}

\DeclareMathOperator*{\argmax}{argmax}

\newcommand*{\argmaxl}{\argmax\limits}

\title{DeePM: A Deep Part-Based Model for Object Detection and Semantic Part Localization}

\author{Jun Zhu, Xianjie Chen \& Alan L. Yuille\\
Department of Statistics\\
University of California, Los Angeles\\
\texttt{\{jzh@,cxj@,yuille@stat.\}ucla.edu} \\
}

\begin{document}
\maketitle

\begin{abstract}
In this paper, we propose a deep part-based model (DeePM) for symbiotic object detection and semantic part localization.
For this purpose, we annotate semantic parts for all 20 object categories on the PASCAL VOC 2012 dataset, which provides information on object pose, occlusion, viewpoint and functionality.
DeePM is a latent graphical model based on the state-of-the-art R-CNN framework, which learns an explicit representation of the object-part configuration with flexible type sharing (e.g., a sideview horse head can be shared by a fully-visible sideview horse and a highly truncated sideview horse with head and neck only).
For comparison, we also present an end-to-end Object-Part (OP) R-CNN which learns an implicit feature representation for jointly mapping an image ROI to the object and part bounding boxes.
We evaluate the proposed methods for both the object and part detection performance on PASCAL VOC 2012, and show that DeePM consistently outperforms OP R-CNN in detecting objects and parts.
In addition, it obtains superior performance to Fast and Faster R-CNNs in object detection.
\end{abstract}

\section{Introduction} \label{sec:intro}
In recent years the use of deep convolutional neural networks (DCNN) \citep{12_nips_krizhevsky,15_iclr_simonyan} has significantly improved the detection of objects.
The region-based convolutional neural network (R-CNN) framework \citep{14_cvpr_girshick, 15_fast_rcnn, 15_faster_rcnn} gave a tremendous performance gain over previous state-of-the-art methods such as the deformable part models (DPMs) \citep{10_TPAMI_felzenszwalb,10_cvpr_zhu_dpm}.
Despite much recent progress in object detection, it is still an open question that how well the DCNN-based methods perform on more complicated vision tasks.
In this paper, instead of simply detecting objects, we are interested in a more challenging task, i.e., \textit{symbiotic objection detection and semantic part localization}.
It requires to detect the objects and localize corresponding semantic parts (if visible) in a unified manner.
To be noted, it is different from the tasks of detecting the objects and parts individually, which do not provide the correspondence between object instances and their parts.

The semantic object parts (e.g., person head, sofa cushion) are of great significance to many vision tasks and deliver important cues for reasoning the object pose, viewpoint, occlusion and other fine-grained properties.
Previous studies involving semantic parts either leverage them to provide more supervision for object detection \citep{12_eccv_azizpour,14_cvpr_chen} or assume that the objects have already been detected \citep{11_cvpr_yang,14_nips_chen}.
In addition, current data annotation of semantic parts either covers only a limited number of articulated object classes (e.g., person, animals) \citep{10_bmvc_johnson,12_eccv_azizpour}, or is not very suitable for detection tasks \citep{14_cvpr_chen}.
To enable a systematic study on our task, we define and annotate the semantic parts for all the $20$ object classes on the PASCAL VOC 2012 dataset.

\begin{figure*}[!t]
\begin{center}
	\includegraphics[width=\linewidth]{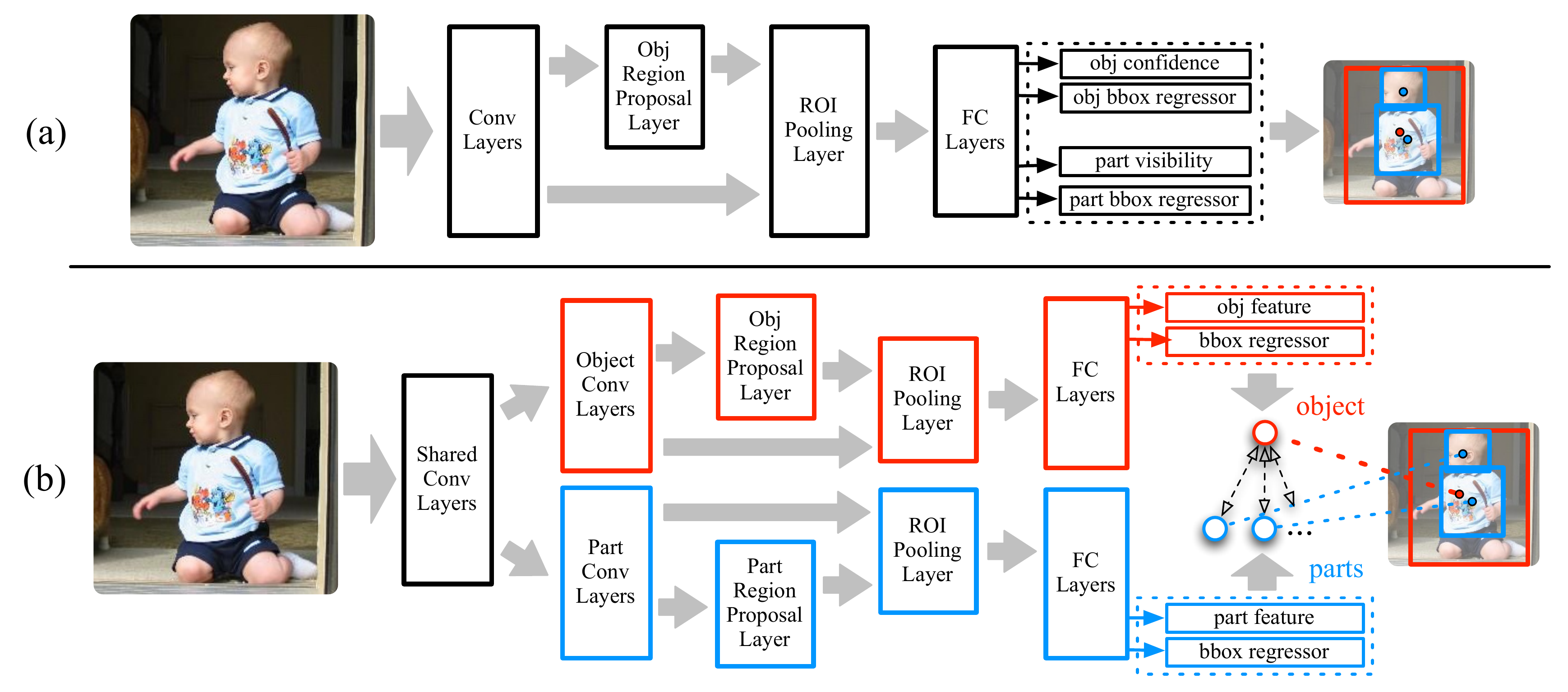}
\end{center}
   \caption{Illustration on the architecture of the proposed models (best viewed in color). (a) the OP R-CNN; (b) the DeePM model.}
\label{fig:model_arch}
\end{figure*}

As shown in Fig.~\ref{fig:model_arch}, on the basis of recently successful R-CNN framework, we explore two different directions of learning representation for our task:
Firstly, we present an end-to-end \textit{Object-Part (OP) R-CNN} (see Fig.~\ref{fig:model_arch} (a)), which learns an implicit deep feature representation for facilitating the mapping from an image ROI to a joint prediction of object and part bounding boxes.
Secondly, we propose a deep part-based model (named \textit{DeePM} in this paper, see Fig.~\ref{fig:model_arch} (b)) which incorporates the Faster R-CNN \citep{15_faster_rcnn} with a part-based graphical model.
It learns an explicit representation on the object-part configuration.

In OP R-CNN, we add two new output layers connected to the last fully-connected layer, and use the corresponding losses for the part visibility classification and bounding-box regression tasks, respectively.
Then, as in Fast \citep{15_fast_rcnn} and Faster \citep{15_faster_rcnn} R-CNNs, we employ a multi-task loss to train the network for joint classification and bounding-box regression on both object and part classes.
DeePM, unlike OP R-CNN, does not directly predict the part location based on the deep feature extracted from the object bounding box.
It adopts a deep CNN with two separate streams, which share the convolutional layers at the early stages and then are dedicated to object and part classes for extracting their appearance features, respectively.
At the same time, as in \citep{15_faster_rcnn}, a region proposal network (RPN) is incorporated in each stream to generate object or part proposals in a learning-based manner.
After that, a part-based graphical model is built to combine the deep appearance features with geometry and co-occurrence constraints between the object and parts.
This enables us to flexibly share part types (learned by unsupervised clustering) and model parameters for learning a compact representation of the object-part configuration.

Using our semantic part annotations on the PASCAL VOC 2012 dataset, we evaluate both object and part detection performance of the proposed methods (OP R-CNN and DeePM), and compare them with the state-of-the-art R-CNN methods on object detection.
The DeePM consistently outperforms OP R-CNN in detecting objects and parts (by $0.3 \%$ and $2.9 \%$ in mAP, respectively), and obtains superior object detection performance to Fast \citep{15_fast_rcnn} and Faster \citep{15_faster_rcnn} R-CNNs.
In addition, we propose a new performance evaluation criterion (named ``$(1+k)$" AP), which considers the detection of object and parts jointly, for the task of symbiotic objection detection and semantic part localization.
The DeePM shows consistently superior performance w.r.t. OP R-CNN in the ``$(1+k)$" AP (when $k>0$), indicating that our flexible graphical model does help to this more complicated detection task.

The rest of this paper is organized as follows:
In Sec.~\ref{sec:related} we discuss the related work of this paper.
Sec.~\ref{sec:data} describes our semantic part annotation on the PASCAL VOC object classes.
In Sec.~\ref{sec:end2endnet} we present the OP R-CNN model.
We elaborate on the DeePM model in Sec.~\ref{sec:model}, and then describe its inference and learning methods in Sec.~\ref{sec:learn}.
We show the experimental results in Sec.~\ref{sec:exp}, and conclude this paper in Sec.~\ref{sec:concl}.

\section{Related Work} \label{sec:related}
The studies of part-based models have a long and important historical standing in computer vision.
The pioneer pictorial structure work \citep{1973_Pictorial_Stucture_Fischler} provided an inspirable framework for representing visual objects with a spring-like graph of parts.
Following this direction, continuous efforts on the part-based models have been made for a wide range of computer vision tasks including object detection \citep{10_TPAMI_felzenszwalb, 14_cvpr_chen}, pose estimation \citep{11_cvpr_yang, 14_nips_chen}, semantic segmentation \citep{15_cvpr_long, 15_iclr_chen} and action recognition \citep{10_cvpr_yang, 13_zhu_iccv_acton}.
Particularly, the DPMs \citep{10_TPAMI_felzenszwalb, 10_cvpr_zhu_dpm}, which are built on basis of the HOG features \citep{05_cvpr_dalal}, have reached milestone of object detection in the past few years.
Different from the ``latent" parts used in \citep{10_TPAMI_felzenszwalb}, some recent works \citep{12_eccv_azizpour, 14_cvpr_chen} have explored the use of semantic parts for improving object detection and localization.
However, semantic part detection has not yet been systemically investigated in literature, which places the main interest of this paper.
Our DeePM model is different from previous part-based object detection methods in the sense of flexible type sharing: Like DPMs \citep{10_TPAMI_felzenszwalb}, the graphical models used in \citep{12_eccv_azizpour, 14_cvpr_chen} are also view-based, where the type of part nodes is tied to the object type.
Compared with other part sharing work \citep{11_cvpr_OttEveringham_sharedparts}, our model enables more flexible sharing between different configurations in the sense that the pairwise edge can be defined on arbitrary pair of object mixture component and part type.
In addition, to be noted, despite the part localization has been successfully addressed in some other tasks such as human pose estimation \citep{11_cvpr_yang, 14_nips_chen}, it is different from our task that the objects are presumably detected beforehand.

Recent advances in visual object recognition are driven by the renaissance of deep convolutional neural networks \citep{98_lecun_gradient, 12_nips_krizhevsky}, which lead to leap progresses in many important recognition tasks such as image classification \citep{12_nips_krizhevsky}, object detection \citep{14_cvpr_girshick} and semantic segmentation \citep{15_cvpr_long, 15_iclr_chen}.
In particular, the line of R-CNN studies has dramatically improved the performance of previous DPMs and become current the state of the art in object detection.
\citep{14_cvpr_girshick} proposed the R-CNN which incorporates the region proposals with DCNN features for object detection task.
\citep{14_eccv_he_spp_net} presented a SPP-net which employs a spatial pyramid pooling layer to efficiently extract the region CNN features.
Very recently, Girshick presented a Fast R-CNN \citep{15_fast_rcnn} which adopts a multi-task loss to enable the joint training of networks for object region classification and bounding-box regression tasks.
In the Faster R-CNN work, \citep{15_faster_rcnn} presented a dedicated RPN, which improves both the runtime and performance by sharing the convolutional layers with Fast R-CNN, to generate object bounding-box proposals efficiently.
In order to enable symbiotic object detection and part localization, the proposed OP R-CNN extends the multi-task loss of Fast/Faster R-CNNs with two additional losses which are responsible for part visibility classification and bounding-box regression.
In our DeePM, the two-stream DCNN, in which the low-level convolutional layers are shared between object and part classes and the latter mid/high-level layers are separate for objects and parts, is dedicatedly designed to our task and naturally conjunct to the graphical model.
In addition, we employ two separate RPNs to generate the region proposals of object and part classes, respectively.

In addition, several researchers \citep{14_savalle_dpm_cnn_feat,15_cvpr_wan,15_cvpr_girshick_dpm_cnn} have married deep CNNs with part-based deformable models, where the parts are not semantic and the part locations are hidden/latent variables.
These models are interesting but their performance is lower than recent R-CNN methods which do not use parts.
\citep{14_eccv_zhang} presented a part-based R-CNN to incorporate semantic part localization with R-CNNs for fine-grained category recognition on birds, which does not address our task in more general objects.
\citep{15_cvpr_Ouyang_DeepIDNet} proposed a DeepID-Net which utilizes a deformation constrained pooling layer to model the deformation of object parts with geometric constraint and penalty in deep convolutional neural networks, but it still uses the ``latent" parts and does not address our task.
\citep{15_cvpr_Zhu_segDeepM} proposed a segDeepM model which incorporates a MRF with the R-CNN to exploit sematic segmentation cues for improving the accuracy of object detection. Although the name of their model is similar to our DeePM, it does not utilize any part annotations and also not involve the task of semantic part detection.


\section{Annotating Semantic Parts for 20 PASCAL VOC Object Classes} \label{sec:data}
On semantic object part annotation, Azizpour and Laptev \citep{12_eccv_azizpour} labelled the part bounding boxes for 6 out of 20 object classes (all are animals) in PASCAL VOC datasets.
In \citep{14_cvpr_chen} it provides pixel-level semantic part annotations for a portion of object classes on PASCAL VOC 2012, but the part definitions of some classes are not suitable for detection tasks (e.g., too small parts like eyes, nose).
Different from these previous works \citep{12_eccv_azizpour,14_cvpr_chen}, we defined and annotated semantically meaningful parts, which are tailored to the task of symbiotic object detection and part localization, for all the 20 object classes in PASCAL VOC 2012 dataset.
Each object category has 1 to 7 parts, resulting in totally 83 part classes.
The definition of parts is based on the body structure (e.g., person, animals), viewpoint (e.g. bus, car) or functionality (e.g., chair, sofa).
Fig. \ref{fig:anno} illustrates our part annotations for the 20 PASCAL VOC object classes.

Our annotation effort took two months of intensely labeling, performed by five labellers trained by us.
This results in much more accurate annotation than using crowdsourcing systems such as MTurk.
For each object instance, the labelers were asked to annotate its visible parts with a tight rectangular bounding box as used in original object-level annotations in PASCAL VOC.
We annotated the parts for all the object instances except some very small or visually difficult ones. In Table \ref{tab:leeds}, we give a full list on our part definition of each object class.
These annotations will be released soon.


\section{The Object-Part R-CNN on Learning Implicit Representation For Object Detection and Part Localization} \label{sec:end2endnet}
In this section, we apply the Faster R-CNN framework \citep{15_faster_rcnn} to the task of symbiotic object detection and part localization.
Besides the original classification and bounding-box regression losses on object classes, we add two new losses which are responsible for the part visibility classification and bounding-box regression tasks, respectively.
Accordingly, there are two additional output layers connected to the last fully-connected (FC) layer, which is shared by all the four sibling output layers for different tasks (see Fig. \ref{fig:model_arch}(a)).
We named it by \textit{OP R-CNN} in this paper.

Suppose there are a total of $J$ object classes and $P_{j}$ part classes for object class $j$.
As the ground-truth annotation used for training our OP R-CNN, each image ROI example is labelled by the following four variables:
an \textit{object class label} $j^{*}$ indicating it belongs to the $j^{*}$-th object ($j^{*} = 0$ corresponds to the background class),
a tuple of \textit{object bounding-box regression targets} $\mathbf{t}_{j}^{*} = (t_{j,\text{x}}^{*}, t_{j,\text{y}}^{*}, t_{j,\text{w}}^{*}, t_{j,\text{h}}^{*})$ for object class $j$ if it is not background,
a \textit{binary part visibility indicator vector} $\mathbf{v}^{*} = (v_{1,1}^{*}, \cdots, v_{1,P_1}^{*}, v_{2,1}^{*}, \cdots, v_{2,P_2}^{*}, \cdots, v_{J,P_J}^{*})$ where $v_{j,i}^{*} = 1 \, (j \in \{1, \cdots, J\} \, \text{and} \, i \in \{1, \cdots, P_j\})$ indicates the visibility of the $i$-th part for the object class $j$,
and a set of \textit{part bounding-box regression targets} $\{ \mathbf{t}_{j,i}^{*} \}$ for all visible parts.

As in \citep{15_fast_rcnn,15_faster_rcnn}, we adopt a multi-task loss $\mathscr{L}$ to train the OP R-CNN for the aforementioned four tasks jointly:
\begin{eqnarray} \label{equ:op_rcnn_loss}
&\mathscr{L}(\mathbf{c}, j^{*}, \mathbf{v}, \mathbf{v}^{*}, \mathbf{t}_{j}, \mathbf{t}_{j}^{*}, \{ \mathbf{t}_{j,i} \}, \{ \mathbf{t}_{j,i}^{*} \}) = \mathscr{L}_{\text{m-cls}}(\mathbf{c}, j^{*}) + \lambda_{o} [j^{*} \geq 1] \mathscr{L}_{\text{loc}}(\mathbf{t}_{j}, \mathbf{t}_{j}^{*}) \nonumber \\
&+ \sum\limits_{j=1}^{J} \sum\limits_{i=1}^{P_{j}} \mathscr{L}_{\text{b-cls}}(v_{j,i}, v_{j,i}^{*})
 + \sum\limits_{j=1}^{J} \sum\limits_{i=1}^{P_{j}} \lambda_{p} [v_{j,i}^{*} = 1] \mathscr{L}_{\text{loc}}( \mathbf{t}_{j,i}, \mathbf{t}_{j,i}^{*} ).
\end{eqnarray}
In Equ. \eqref{equ:op_rcnn_loss}, $\mathbf{c} = (c_0, c_1, \cdots, c_J)$ is the predicted object class probability vector, which is computed by the softmax operation over the $(J+1)$ confident values from the object classification output layer.
$v_{j,i}$ is the predicted probability on the presence of the $i$-th part of the object class $j$.
$\mathscr{L}_{\text{m-cls}}(\mathbf{c}, j^{*}) = -\log(c_{j^{*}})$ is a log loss for multi-class object classification task, while $\mathscr{L}_{\text{b-cls}}(v_{j,i}, v_{j,i}^*) = v_{j,i}^{*} \log(v_{j,i}) + (1-v_{j,i}^{*})\log(1-v_{j,i})$ is a binary cross-entropy loss for part visibility classification task.
Following \citep{15_fast_rcnn}, $\mathscr{L}_{\text{loc}}(\mathbf{t}, \mathbf{t}^{*}) = \sum_{u \in \{\text{x}, \text{y}, \text{w}, \text{h}\}} \text{smooth}_{L_1} (t_{u} - t_{u}^*)$ (Here we omit the class index subscript $j$ or $j,i$ for notation brevity) is a smooth $L_1$ loss for bounding-box regression task.
$[\cdot]$ is the Iverson bracket indicator function which outputs 1 when the involved statement is true and 0 otherwise.
It implies that we only use the positive examples (i.e., in case that the ground-truth bounding box is viable) for training the bounding-box regressors.

\section{The DeePM Model On Explicit Representation of Object and Semantic Parts} \label{sec:model}
As shown in Fig. \ref{fig:model_arch}(b), our DeePM is a detection pipeline composing of a two-stream DCNN and a latent graphical model.
This DCNN is dedicatedly designed for our task (symbiotic object detection and part localization).
It is responsible for generating both of the object and part proposals as well as corresponding deep features which are used in the appearance terms of the graphical model.
The graphical model is presented to incorporate the deep features with the object-part geometry and co-occurrence constrains, which are dependent on different object and part types to capture typical configurations.
We will describe the details of the DCNN and the graphical model in Sec.~\ref{ssec:faster_rcnn} and \ref{ssec:gpm}, respectively.


\subsection{A Two-Stream DCNN for Objects and Parts} \label{ssec:faster_rcnn}
We propose a two-stream DCNN to generate detection proposals and appearance features for both the object and semantic parts based on the Faster R-CNN framework.
Its architecture is illustrated in Fig.~\ref{fig:model_arch} (b).
The convolutional layers are shared in the early stages, and then split into two separate streams which correspond to object and part classes, respectively.
This is desirable because the low-level visual representations (e.g., oriented edges, color blobs) are commonly shared among different classes but the mid-level ones should be class-specific.
After that, all the subsequent layers are designed for objects and parts in separate streams.
For either the object-level or part-level stream, we adopt a RPN \citep{15_faster_rcnn} to generate the object or part bounding-box proposals correspondingly.
The RPNs, which share the convolutional layers with the detection networks, can efficiently generate the object and part proposals in a learning-based manner.
It is more desirable (especially for parts) than traditional region proposal methods (e.g., Selective Search \citep{13_ijcv_uijlings_selective_search}) which are usually based on low-level or middle-level visual cues.

In each stream, the rectangular detection proposals are manufactured with a ROI pooling layer~\citep{14_eccv_he_spp_net,15_fast_rcnn}.
It generates a fixed dimensional representation based on the feature activities of the last convolutional layer.
The pooled features of detection proposals are then fed into the fully-connected (FC) layers, followed by a bounding-box regression layer for improving the localization accuracy.
We use the last FC layer's output activities as the feature representation of the appearance terms in the subsequent graphical model (see Fig. \ref{fig:model_arch}(b)).

In training the part-level stream of our DCNN, we combine some part classes together into one category.
For instance, some parts, which are defined in the sense of spatial positions w.r.t. the object center, have indistinguishable visual appearance (e.g., front and back bicycle wheels) so they should be merged into one category for training the deep network.

\subsection{A Part-Based Graphical Model with Flexible Sharing} \label{ssec:gpm}
For an object class, we define a constellation model with $(P+1)$ nodes where $P$\footnote{In this section, for notation brevity, we omit the object class index $j$ used in Sec.~\ref{sec:end2endnet}.} is the number of semantic parts involved in this object.
Let $i = 0$ denote the object node and $i \in \mathcal{P} = \{1, \cdots, P\}$ index the node of the $i$-th part.
For each node $i$, we parameterize its location in the image $\mathbf{I}$ by a bounding box $\mathbf{l}_i = (l^{x}_{i}, l^{y}_{i}, l^{w}_{i}, l^{h}_{i})$.
To account for the visual variations of an object or a part, we introduce a ``type" variable $z_i \in \mathcal{K}_i$ for each node $i$.
Let $\mathcal{K}_0 = \{1, \cdots, K_0\}$ and $\mathcal{K}_i = \{0, 1, \cdots K_i\}$ ($i \neq 0$) denote the candidate type sets of the object and the $i$-th part, respectively.
Particularly, we use the type value $z_i = 0$ to represent the invisibility of the part $i$.
Then we define the configuration of an object associated with its parts by $(\mathbf{L}, \mathbf{Z})$, where $\mathbf{L} = (\mathbf{l}_0, \mathbf{l}_1, \cdots, \mathbf{l}_{P})$ and $\mathbf{Z} = (z_0, z_1, \cdots, z_{P})$.

In our graphical model there are three kinds of terms: the appearance term $S^{a}$, the geometry compatibility term $S^{g}$ and the bias term $S^{b}$.
We define the scoring function of a configuration of the object and its parts as Equ. \eqref{equ:score_func}:
\begin{equation} \label{equ:score_func}
S(\mathbf{I}, \mathbf{L}, \mathbf{Z}) = \sum_{i=0}^{P} S_{i}^{a}(\mathbf{I}, \mathbf{l}_i, z_i) + \sum_{i=1}^{P} S_{i}^{g}(\mathbf{I}, \mathbf{l}_i, z_i, \mathbf{l}_0, z_0) + \sum_{i=1}^{P} S_{i}^{b}(z_i,z_0) + S_{0}^{b}(z_0).
\end{equation}

{\bf The appearance term:}
The appearance term of each node is defined as a linear model $\mathbf{w}^{a}_{i,z_i}$ on the last FC layer's feature.
For a bounding box $\mathbf{l}_i$ on $\mathbf{I}$, let $\boldsymbol{\phi}^{o}(\mathbf{I},\mathbf{l}_i)$ and $\boldsymbol{\phi}^{p}(\mathbf{I},\mathbf{l}_i)$ denote the last FC layer's features of our DCNN's object and part streams, respectively.
Formally, it is defined by $S_{0}^{a}(\mathbf{I}, \mathbf{l}_0, z_0) = \mathbf{w}^{a}_{0,z_0} \boldsymbol{\phi}^{o}(\mathbf{I},\mathbf{l}_0)$ (for the object node) or $S_{i}^{a}(\mathbf{I}, \mathbf{l}_i, z_i) = \mathbf{w}^{a}_{i,z_i} \boldsymbol{\phi}^{p}(\mathbf{I},\mathbf{l}_i)$ (for the part node $i$).
We can see that the model parameter $\mathbf{w}^{a}_{i,z_i}$ depends on the object or part type $z_i$ which accounts for typical part configurations or visual variations correspondingly.

{\bf The geometry compatibility term:}
Similar to \citep{14_cvpr_chen}, we use a vector $\mathbf{d}_i = [dx_{i}, dy_{i}, sx_{i}, sy_{i}, s_{i}]$ to represent the spatial deformation of the part $i$ w.r.t. the object, where $dx_{i} = \frac{l^{x}_{i} - l^{x}_{0}}{l^{w}_{i} + l^{w}_{0}}$, $dy_{i} = \frac{l^{y}_{i} - l^{y}_{0}}{l^{h}_{i} + l^{h}_{0}}$ are the normalized spatial displacements and $sx_{i} = \frac{l^{w}_{i}}{l^{w}_{0}}$, $sy_{i} = \frac{l^{h}_{i}}{l^{h}_{0}}$, $s_{i} = \sqrt{sx_{i} \cdot sy_{i}}$ are the normalized scales.
Furthermore, we define a type-specific prototype parameter of the geometry term by $\boldsymbol{\mu}_{i,z_i,z_0} = [\mu^{dx}_{i,z_i,z_0}, \mu^{dy}_{i,z_i,z_0}, \mu^{sx}_{i,z_i,z_0}, \mu^{sy}_{i,z_i,z_0}, \mu^{s}_{i,z_i,z_0}]$, which specifies the ``anchor" point in the geometry feature space.
Thus the geometry term $S_{i}^{g}(\mathbf{I}, \mathbf{l}_i, z_i, \mathbf{l}_0, z_0) = \mathbf{w}^{g}_{i,z_i,z_0} \boldsymbol{\psi}(\mathbf{d}_i; \boldsymbol{\mu}_{i,z_i,z_0})$ measures the geometry compatibility between the object and the part $i$, where $\mathbf{w}^{g}_{i,z_i,z_0}$ is a type-specific weight vector.
Meanwhile, $\boldsymbol{\psi}(\mathbf{d}_i; \boldsymbol{\mu}_{i,z_i,z_0})$ is a feature vector linearizing the quadratic deformation penalty of $\mathbf{d}_i$ w.r.t. $\boldsymbol{\mu}_{i,z_i,z_0}$, i.e., $\boldsymbol{\psi}(\mathbf{d}_i; \boldsymbol{\mu}_{i,z_i,z_0}) = - [(dx_{i} - \mu^{dx}_{i,z_i,z_0})^{2}, (dx_{i} - \mu^{dx}_{i,z_i,z_0}), \nonumber (dy_{i} - \mu^{dy}_{i,z_i,z_0})^{2}, (dy_{i} - \mu^{dy}_{i,z_i,z_0}), (sx_{i} - \mu^{sx}_{i,z_i,z_0})^{2}, (sx_{i} - \mu^{sx}_{i,z_i,z_0}), (sy_{i} - \mu^{sy}_{i,z_i,z_0})^{2}, (sy_{i} - \mu^{sy}_{i,z_i,z_0}), (s_{i}  - \mu^{s}_{i,z_i,z_0})^{2},   (s_{i} - \mu^{s}_{i,z_i,z_0})]^{\top}$.

{\bf The bias term:}
We define bias terms to model the prior belief of different object-part configurations and type co-occurrence.
Specifically, the unary bias term $S_{0}^{b}(z_0) = b^{0}_{z_0}$ favors particular type assignments for the object node, while the pairwise bias term $S_{i}^{b}(z_i,z_0) = b^{i}_{z_i,z_0}$ favors particular type co-occurrence patterns between the object and the part $i$.
\begin{figure*}[!t]
\begin{center}
	\includegraphics[width=0.8\linewidth]{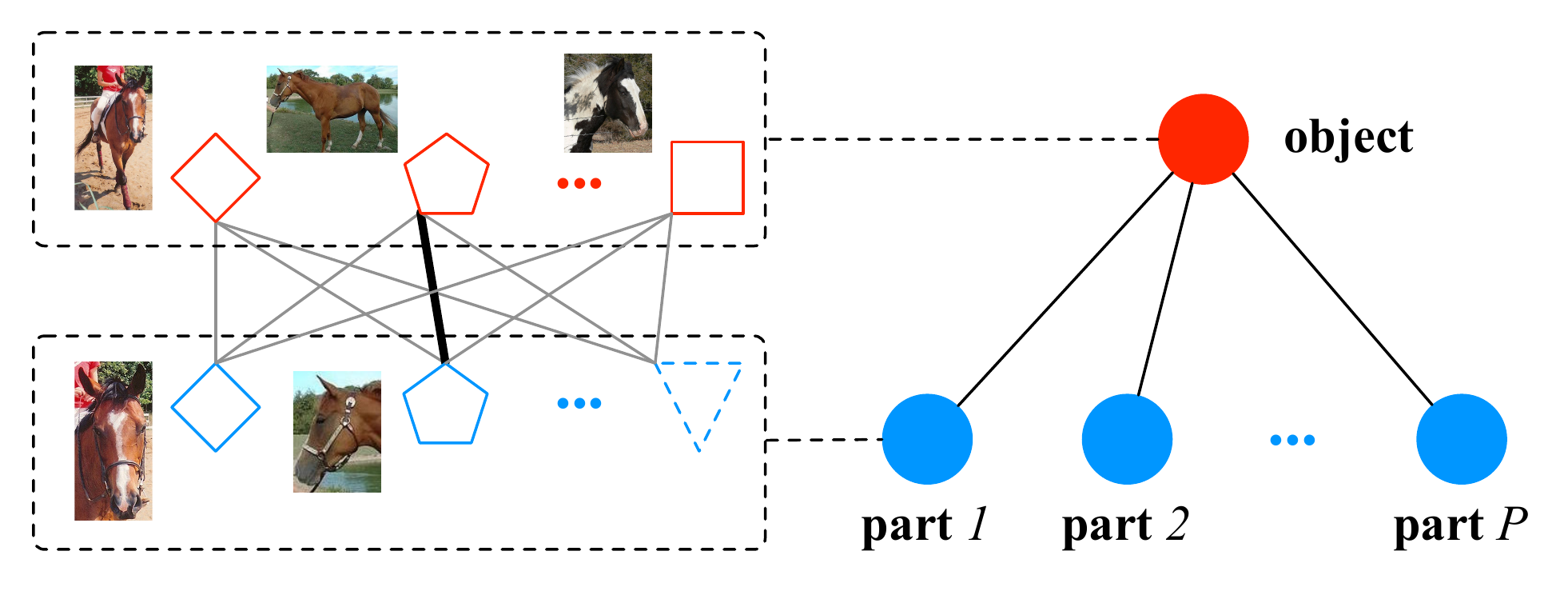}
\end{center}
   \caption{Illustration on flexible type sharing in DeePM. The red solid circle represents the object node, and the blue solid circles indicate the part nodes. The red and blue hollow marks correspond to different types of object and part nodes, respectively. In our graphical model, one part type (i.e., a sideview horse head) could be shared by different object types (i.e., a fully-visible sideview horse, a highly occluded horse with only head and neck). In the inference step, only one type-specific pairwise relation (i.e., the bold edge) would be selected from the candidate type pair set for each occurrence or geometric term (best viewed in color).}
\label{fig:type_sharing}
\end{figure*}

{\bf Flexible part type sharing:}
In this paper, the object type represents global object-part configuration (e.g., a particular object pose or viewpoint), while the part type corresponds to a typical visual appearance component in a mixture distribution.
In our DeePM model, the pairwise edges allow the connections on arbitrary object-part type pairs for each part $i$.
This enables flexible part type sharing among different object configurations, which is different from the tied object-part types in previous DPM models \citep{10_TPAMI_felzenszwalb,12_eccv_azizpour,14_cvpr_chen} on object detection.
For example, as illustrated in Fig. \ref{fig:type_sharing}, one part type (i.e., a sideview horse head) could be shared by different horse object types (i.e., a fully-visible sideview horse, a highly occluded horse with only head and neck).
This is desirable in the sense of compactness and efficiency on representation.

{\bf Flexible model parameter sharing:}
As mentioned in Sec. \ref{ssec:faster_rcnn}, some part classes are merged to one category in training the DCNN, making the appearance features non-discriminative among these parts.
Accordingly, this requires the model parameter of their appearance terms ought to be shared.
E.g., as shown in Fig. \ref{fig:type_sharing}, the appearance model of the sideview horse head is invariant to different orientations in the horizontal direction (i.e., towards left and right).

In this paper, DeePM is formulated in a more flexible and modular manner than previous DPM models.
For this purpose we introduce a dictionary of part appearance models denoted by $\mathcal{D}^{a}_{\mathbf{w}} = \{ \mathbf{w}^{a}_{1}, \cdots, \mathbf{w}^{a}_{M_{a}} \}$.
$\mathbf{w}^{a}_{k}$ ($k \in \{1, \cdots, M_{a}\}$) is an elemental appearance model which could be shared by different parts.
Here we adopt the same notation $\mathbf{w}^{a}$ as before to indicate the appearance model parameter, but use a subscript of linear index $k$ instead.
Then we define an index mapping $\rho_{a}(\cdot,\cdot): (P, K_i) \rightarrow M_{a}$ to transform the two-tuple subscript $(i,z_i)$ to $k$.
$\rho_{a}$ is generally a many-to-one mapping function which enables flexible sharing of the appearance models between different parts (even the sharing could be allowed for the parts from different object classes).
Likewise, we also define a dictionary of geometry models $\mathcal{D}^{g}_{\mathbf{w}} = \{ (\mathbf{w}^{g}_{1}, \boldsymbol{\mu}_{1}), \cdots, (\mathbf{w}^{g}_{M_{g}}, \boldsymbol{\mu}_{M_{g}}) \}$ and corresponding index mapping function $\rho_{g}(i,z_i,z_0)$, where $\boldsymbol{\mu}_{k}$ ($k \in \{1, \cdots, M_{g}\}$) is the geometry prototype associated with the $k$-th element $\mathbf{w}^{g}_{k}$ in the dictionary.

Now we rewrite the model scoring function (Equ. \eqref{equ:score_func}) as below:
\begin{eqnarray} \label{equ:score_func_alt}
S(\mathbf{I}, \mathbf{L}, \mathbf{Z}) = {\mathbf{w}^{a}_{0,z_0}} \boldsymbol{\phi}^{o}(\mathbf{I}, \mathbf{l}_0)
                      + \sum_{k=1}^{M_{a}} {\mathbf{w}^{a}_{k}} [\sum_{\rho_{a}(i,z_i)=k} \boldsymbol{\phi}^{p}(\mathbf{I},\mathbf{l}_i)] \nonumber\\
                      + \sum_{k'=1}^{M_{g}} {\mathbf{w}^{g}_{k'}} [\sum_{\rho_{g}(i,z_i,z_0)=k'} \boldsymbol{\psi}(\mathbf{d}_i; \boldsymbol{\mu}_{k'})]
                      + b^{0}_{z_0} + \sum_{i=1}^{P} b^{i}_{z_i,z_0}.
\end{eqnarray}

Generally the bias parameters should not be shared between different types because it may make the type prior non-informative.
In practice, we manually specify $\rho_{a}$ based on the appearance similarity of the parts, which is consistent to the merged part category definition in training the DCNN.
For $\rho_{g}$, we do not impose model parameter sharing on the geometry terms.
To be noted, one advantage of our formulation is that the sharing of model parameters can be specified in part type level and even be learned automatically (i.e., learning the mapping function $\rho$ on the fly).
We will explore this direction in future work.

\section{Inference and Learning on DeePM} \label{sec:learn}
As mentioned in Sec. \ref{ssec:faster_rcnn}, we generate the object and part detection proposals via corresponding object-level and part-level RPNs, respectively.
Let $\mathcal{B}^{o}_{\mathbf{I}}$ denote an object proposal set and $\mathcal{B}^{p}_{\mathbf{I}}$ be a part proposal set for $\mathbf{I}$.
For any object bounding box $\mathbf{l}_0 \in \mathcal{B}^{o}_{\mathbf{I}}$, we define the set of its candidate part windows by $\mathcal{B}^{p}_{\mathbf{I}}(\mathbf{l}_0,\tau_{in}) = \{\mathbf{l} | \mathbf{l} \in \mathcal{B}^{p}_{\mathbf{I}} \; \text{and} \; in(\mathbf{l},\mathbf{l}_0) \geq \tau_{in} \}$, where $in(\mathbf{l},\mathbf{l}_0)$ is the inside rate measuring the area fraction of a window $\mathbf{l}$ inside $\mathbf{l}_0$ and $\tau_{in}$ is a threshold ($\tau_{in} = 0.8$ in this paper).
Thus given $\mathbf{l}_0$ the best configuration $(\hat{\mathbf{L}}, \hat{\mathbf{Z}})$ can be inferred by maximizing the scoring function:
\begin{equation} \label{equ:inference}
(\hat{\mathbf{L}}, \hat{\mathbf{Z}}) = \argmaxl_{\forall i \in \mathcal{P}, \, \mathbf{l}_i \in \mathcal{B}^{p}_{\mathbf{I}}(\mathbf{l}_0,\tau_{in}); \, \forall i, \, z_{i} \in \mathcal{K}_i} S(\mathbf{I}, \mathbf{L}, \mathbf{Z}).
\end{equation}
We use dynamic programming to maximize Equ.~\eqref{equ:inference} in inference.
In the rest of this section, we elaborate on the learning method of our DeePM.

{\bf Learning the DCNN:}
Generally we use similar criterions to train the DCNN as in \citep{15_faster_rcnn}.
The shared convolutional layers are directly inherited from a pre-trained network.
The fine-tuning procedure starts from the separate convolutional layers throughout all subsequent layers in both the object-level and part-level streams.
To enable the sharing of convolutional layers for RPN and Fast R-CNN, we follow the four-step stage-wise training criterion as in \citep{15_faster_rcnn}.

{\bf Learning object and part types:}
In this paper, the object types are defined to capture typical configurations while the part types account for different components in a mixture of visual appearances. Thus we learn the types by using two different criterions for objects and parts, respectively.

Similar to \citep{12_eccv_azizpour}, we adopt a pose-based global configuration feature to learn the object types.
Concretely, for each positive ground-truth example we concatenate the geometry features $\{\mathbf{d}_i\}$ of all visible parts as well as the binary part visibility indicators to a single vector, and then use it as the global configuration feature to perform a modified K-means clustering algorithm \citep{12_eccv_azizpour} which can well handle missing data (i.e., some parts may be absent)\footnote{Please refer to \citep{12_eccv_azizpour} for the details of this modified K-means clustering algorithm.}.
After that, we obtain a couple of clusters as object types which potentially correspond to different typical configurations.
Fig. \ref{fig:vis_type_clusters_horse} and \ref{fig:vis_type_clusters_person} visualize the learned types for horse and person, respectively.

For part classes, we adopt the feature activities after the ROI pooling layer to learn the types.
Specifically such features from the DCNN's part-level stream are used to perform K-means clustering on the ground-truth part data, and the resultant clusters work as part types.

{\bf Learning the graphical model with latent SVM:}
To facilitate the introduction of learning our graphical model, we gather all the model parameters $\mathbf{w}$ and $b$ into one single vector $\mathbf{W} = [\mathbf{w}^{a}_{0,1}, \cdots, \mathbf{w}^{a}_{0,K_0}, \mathbf{w}^{a}_{1}, \cdots, \mathbf{w}^{a}_{M_a}, \mathbf{w}^{g}_{1}, \cdots, \mathbf{w}^{g}_{M_g}, b^{0}_{1}, \cdots, b^{0}_{K_0}, b^{1}_{0,1}, \cdots, b^{1}_{K_1,1}, b^{2}_{0,2}, \cdots, b^{P}_{K_{P},K_0}]^{\text{T}}$.
Given labelled positive training examples $\{(\mathbf{I}_n, \mathbf{L}_n)\}$ ($y_n = 1$) and negative examples $\{(\mathbf{I}_n)\}$ ($y_n = -1$), where $\mathbf{L}_n$ involves all the visible ground-truth part bounding boxes for the example $n$ (the negative examples do not have parts), we learn the model parameters $\mathbf{W}$ via the latent SVM framework \citep{10_TPAMI_felzenszwalb}:
\begin{eqnarray} \label{equ:latent_svm}
&\min_{(\mathbf{W}, \, \xi_n \geq 0)} \frac{1}{2} \mathbf{W}^{\text{T}} \mathbf{W} + C \sum_{n} \xi_n \nonumber\\
\text{s.t.} \; \forall \, n \,\, \text{and} \,\, (\mathbf{L},\mathbf{Z}) \in \mathcal{H}_n, \;\; &y_n \cdot \max_{(\mathbf{L},\mathbf{Z})} S(\mathbf{I}_n, \mathbf{L}, \mathbf{Z}; \mathbf{W}) \geq 1 - \xi_n ,
\end{eqnarray}
where $C$ is a predefined hyper-parameter on model regularization and $\mathcal{H}_n$ represents a feasible set of latent configurations for example $n$.
In this paper, the definition of $\mathcal{H}$ is different between positive and negative examples:
For positive examples, we constrain the search space $\mathcal{H}_n$ with the ground-truth bounding boxes of the parts which are visible (i.e., the candidate part locations $\mathbf{L}$ should be consistent to $\mathbf{L}_n$, and the candidate type values of $z_i$ should be larger than 0 for any visible part $i$).
This constraint encourages the correct configurations of positive examples to be scored higher than a positive margin value (i.e., $+ 1$).
For negative examples, in contrast, we do not impose any restrictions on $\mathcal{H}_n$, implying that the scores should be less than a negative margin (i.e., $-1$) for all possible configurations of part locations and types.

Due to the existence of latent variables for positive examples, the problem of Equ. \eqref{equ:latent_svm} is not convex and thus we employ the CCCP algorithm \citep{yuille2002concave} to minimize the loss iteratively.
At first, we initialize the object and part type values of positive examples according to the assignments from the aforementioned type clustering stage.
Then we iteratively optimize Equ. \eqref{equ:latent_svm} by alternating between two steps:
(1) Given the type value assignments of positive examples, Equ. \eqref{equ:latent_svm} becomes convex and we use a dual coordinate quadratic program solver \citep{13_dcs_svm_solver} to minimize the hinge loss.
(2) Given current model $\mathbf{W}$, we search the best type assignments of the object and visible parts for positive examples. By iterations of these two steps, the loss of \eqref{equ:latent_svm} decreases monotonously till convergence.




\section{Experiments} \label{sec:exp}
\subsection{The Visual Task and Evaluation Criterion} \label{ssec:task_eval_criterion}
We evaluate the proposed OP R-CNN and DeePM models on the task of symbiotic object detection and semantic part localization.
It requires the model to output all detected object bounding boxes with corresponding part bounding boxes (if visible).
Each object bounding box is associated with a prediction score or probability indicating the confidence of presence on the object class of interest.
Likewise, all the output part bounding boxes are associated with the visibility confidence values of corresponding part classes.
To be noted, our task is different and more challenging than independent object or part detection in the sense that it asks for the correspondence between the output object and part detections. E.g., it requires to know which head bounding box corresponds to some particular person bounding box in a couple of overlapping person detections.

In this paper, we first use the \textit{average precision} (AP) \citep{10_IJCV_Everingham_PASCAL_VOC}, which is a standard evaluation criterion used in the object detection literature, as the performance evaluation criterion for both the object and part detection tasks.
Particularly, it is a stricter measurement on evaluating the part localization performance than the \textit{percentage of correct part} (PCP) criterion which is commonly used in the pose estimation literature.
PCP only considers the part detections involved in true positive object bounding boxes, making it less informative for false positive detections.
Because we cannot assume that the object bounding boxes are perfectly detected in advance of part localization in our task, AP is a more suitable performance evaluation criterion than PCP for part detection.

However, because it calculates the object and part detection performance separately, the standard AP criterion does not suit for the task of symbiotic objection detection and semantic part localization.
In this paper, we propose a new performance evaluation criterion (named ``$(1+k)$" AP) for this task.
Specifically, the ``$(1+k)$" AP is defined as the average precision in the sense that both the object and at least $k$ parts of it are correctly detected (i.e., the IoU overlap w.r.t. ground-truth object/part bounding box is larger than 0.5, or no bounding box output for invisible parts).
For instance, the ``$(1+2)$" AP means that only the detections, in which both of the object and no less than 2 parts are predicted correctly, are regarded as true positive examples.
Like the standard PASCAL VOC AP criterion \citep{10_IJCV_Everingham_PASCAL_VOC}, the duplicate detections are punitively counted as false positives.
Thus, the proposed ``$(1+k)$" AP criterion would produce a brunch of AP numbers, each of which corresponds to a different number requirement of parts correctly detected.
For $k=0$, it does not require to detect parts correctly and this AP number is corresponding to the performance of detecting objects solely.
When $k$ is equal to the number of all possible parts for an object class, the ``$(1+k)$" AP number will be the most strict one because only the perfect joint object-part detections (i.e., both the object and all the corresponding parts are correctly detected) can be counted as true positive examples.

\subsection{Experimental Settings} \label{ssec:exp_setting}
Because we only annotate the semantic parts on the \textit{trainval} images of PASCAL VOC 2012, we first perform several diagnostic experiments as well as part detection evaluation by using \textit{train} set for training and \textit{val} set for testing on the diagnostic experiments.
In addition, we test our methods for the object detection task with VOC 2012 \textit{test} set, and compare the object detection performance with Faster R-CNN \citep{15_faster_rcnn}.
The AP number is evaluated for each (either object or part) class individually and the mean AP (mAP) is reported.

We construct our OP R-CNN based on the Faster R-CNN architecture \citep{15_faster_rcnn}, in which it uses the RPN to generate region proposals.
Similar to \citep{15_fast_rcnn, 15_faster_rcnn}, we set the loss balance hyper-parameters by $\lambda_{o} = 1$ and $\lambda_{p} = 1$.
All other parameters in training the OP R-CNN follow the settings in \citep{15_faster_rcnn}.
All the DCNNs in our experiments are fine-tuned from a pretrained VGG-16 net \citep{15_iclr_simonyan}.

In training the DCNN for our DeePM model, we follow the parameter settings in \citep{15_faster_rcnn}.
The convolutional layers from $\textit{conv}1\_1$ to $\textit{conv}2\_2$ are shared in the DCNN, and the separate streams start from $\textit{conv}3\_1$.
We use the feature activities of the last FC layer of VGG-16 net (i.e., $\textit{fc}7$) as the appearance feature in the graphical model.
We also normalize the $\textit{fc}7$ features as in \citep{14_cvpr_girshick}, and set $C = 0.001$.
Similar to \citep{14_cvpr_girshick}, we use hard negative example mining over all images, with the IoU overlap threshold $0.3$ (the object proposal windows with overlap less than $0.3$ w.r.t. the ground-truth boxes are used as negative examples).
We use $5$ types for each object class and $3$ types for each part class.
In testing stage, we use the same inference heuristics as in \citep{15_faster_rcnn}.
The bounding-box regressors, which are learned from the object-level and part-level streams of the DCNN, are used in the non-maximum suppression (NMS) operations of object and part detections, respectively.
For comparison, we obtain the Fast R-CNN and Faster R-CNNs' results by using the code released from the authors of \citep{15_faster_rcnn}.
For Fast R-CNN \citep{15_fast_rcnn}, we use the selective search approach \citep{13_ijcv_uijlings_selective_search} to generate object region proposals.

\subsection{Experiment Results and Analysis} \label{ssec:exp_result}
We first performance diagnostics experiments on OP R-CNN and DeePM, and then compare them with other object detection methods.

Table~\ref{tab:perf_obj_det} shows the detection performance for object classes.
In this experiment, we conduct a baseline method which uses the same $\textit{fc}7$ features of the DCNN in DeePM to train SVM classifiers for object detection.
This enables a direct comparison with our DeePM, in order to investigate the significance of using the geometry and co-occurrence cues in the graphical model.
We use `$\textit{fc}7+$svm' to denote this baseline method in table~\ref{tab:perf_obj_det}.
The DeePM outperforms the $\textit{fc}7+$svm baseline by $1.4 \%$ in mAP, showing the significance of the geometry and co-occurrence constraints in the explicit representation learned with semantic parts.
Moreover, its performance is comparable with OP R-CNN and superior to Fast R-CNN and Faster R-CNN on object detection.

For part detection, as shown in table~\ref{tab:perf_part_det}, the DeePM model shows superior performance ($2.9 \%$ in mAP) w.r.t. OP R-CNN.
Especially, the performance improvement tends to be relatively large for small parts (e.g., the animal tails, heads).
OP R-CNN learns an implicit feature representation extracted from the object bounding box to regress the part location, which may be difficult to predict potentially very variational positions for small parts.
By contract, DeePM employs the part-level stream of DCNN to generate part bounding-box proposals, and then explicitly leverages useful geometry and co-occurrence cues to localize the parts in a symbiotic manner.

Moreover, we compare the DeePM with a deformable part-based baseline model, which has a single object type and the same set of semantic parts each with a single type. This baseline DPM model is basically a `DPM-v1' counterpart \citep{10_TPAMI_felzenszwalb} on top of the learned deep features and RPN proposals.
In figure~\ref{fig:compr_single_type_model}, we show the performance comparison between DeePM (5 types used for the object node, 3 types used for the part nodes) and the `DPM-v1'-like single type baseline model on four object classes, i.e. bicycle, boat, horse and sofa.
When using only a single type for each node, the geometric and co-occurrence terms will be not informative, leading to poor performance of detecting the object and its parts.
As shown in figure~\ref{fig:compr_single_type_model}, the DeePM model generally outperforms the single-type `DPM-v1' baseline for both object and part detection performance, indicating the significance of using type-specific geometric and co-occurrence cues in the graphical model.
In addition, following the strategy proposed in \citep{12_eccv_hoiem}, we give a detailed analysis on the performance w.r.t. object/part size in figure~\ref{fig:compr_perf_wrt_size}.
We can see that the extremely small (`XS') or small (`S') instances are very difficult to be detected, and the performance of extremely large (`XL') objects or parts is also relatively low because of highly truncated/occuluded examples often occurred in this size level.

The experiments above evaluate the object and part detection performance separately, which enable us to compare our methods with previous object detection approaches and show detection performance for each individual part class.
However, as mentioned in Sec.~\ref{ssec:task_eval_criterion}, the standard evaluation criterion (i.e., PASCAL VOC AP criterion) does not suit for our main task (i.e., symbiotic objection detection and semantic part localization) in this paper.
For this purpose, we also test the joint object-part detection performance for these four object classes by using the proposed ``$(1+k)$" AP criterion.
As shown in figure~\ref{fig:compr_one_plus_k_aps}, we can see that the performance decreases dramatically when the quantity requirement of correctly detected parts (i.e., $k$) becomes larger, which implies that corresponding detection task is more difficult.
For the situation that $k$ is equal to the number of all possible parts, the ``$(1+k)$" AP is extraordinary low (usually less than $1 \%$) caused from the extremely strict definition of true positive examples.
The DeePM consistently outperforms OP R-CNN\footnote{For OP R-CNN, the part detections will be regarded as the invisible ones if the corresponding visibility probabilities are lower than 0.5.} in this ``$(1+k)$" AP (when $k>0$), which indicates that our flexible graphical model does help to a more complicated detection task like the joint detection of object and semantic parts.

At last, we test the object performance of our methods on VOC 2012 \textit{test} set.
In this experiment, the models are trained with VOC 2012 \textit{trainval} set and the parameter settings are consistent with above diagnostic experiments.
As shown in table~\ref{tab:perf_obj_det_voc12test}, we can see that the OP R-CNN and DeePM obtain slightly superior performance w.r.t. Fast/Faster R-CNN.
In figure~\ref{fig:vis_det}, we visualize some examples of the DeePM's detection result.

\section{Conclusion} \label{sec:concl}
In this paper, we study on learning part-based representation for symbiotic object detection and semantic part localization.
For this purpose, we annotate semantic parts for all the 20 object classes on the PASCAL VOC 2012 dataset, which provides information on reasoning object pose, occlusion, viewpoint and functionality.
To deal with, we propose both implicit (OP R-CNN) and explicit (DeePM) solutions.
We evaluate our methods for both the object and part detection on PASCAL VOC 2012, and show that DeePM consistently outperforms OP R-CNN (especially by a relatively large margin on part detection), implying the importance of using the learning-based part proposals and explicit geometry cues for part localization.
In addition, we proposed a new ``$(1+k)$" AP performance criterion for evaluating the task of symbiotic objection detection and semantic part localization.
The DeePM consistently outperforms OP R-CNN in the ``$(1+k)$" AP (when $k>0$), indicating that our flexible graphical model does help to this more complicated detection task.


\bibliography{iclr2016_conference}
\bibliographystyle{iclr2016_conference}

\newpage
\appendix
\section{Supplementary} \label{sec:supp}

\begin{table} [!bp]
\centering
\rowcolors{2}{}{gray!35}
\addtolength{\tabcolsep}{-4.0pt}
\scriptsize
\begin{tabular}{ c | c | c c c c c c c c c c c c c c c c c c c c }
\toprule[0.2 em] %
 Method          &  mAP & areo &  bike &  bird &  boat &  bottle &  bus &  car &  cat &  chair &  cow &  table &  dog &  horse &  mbike &  person &  plant &  sheep &  sofa &  train &  tv \\
\textit{fc}7+svm & 65.2 & 78.2 & 73.3 & 66.4 & 46.0 & 41.9 & 78.6 & 69.1 & 85.8 & 42.4 & 63.7 & 50.3 & 83.0 & 71.2 & 76.8 & 76.6 & 34.2 & 68.8 & 58.1 & 76.8 & 62.6 \\
\textit{fast}    & 64.7 & 79.9 & 74.1 & 67.3 & 45.6 & 35.1 & 77.0 & 64.8 & 86.2 & 37.0 & 65.4 & 53.0 & 83.6 & 74.8 & 77.6 & 69.8 & 31.4 & 68.6 & 59.7 & 77.3 & 66.2 \\
\textit{faster}  & 65.6 & 79.6 & 74.7 & 68.5 & 47.7 & 41.3 & 79.7 & 69.7 & 86.6 & 39.9 & 63.7 & 50.4 & 83.5 & 73.9 & 78.6 & 77.0 & 33.0 & 69.8 & 56.6 & 76.6 & 62.0 \\
\midrule
\textit{op rcnn} & 66.3 & 79.6 & 76.0 & 67.4 & 48.7 & 42.6 & 80.6 & 69.9 & 86.5 & 40.5 & 67.0 & 52.8 & 84.2 & 75.3 & 78.2 & 77.8 & 31.4 & 68.8 & 57.8 & 80.4 & 60.8 \\
DeePM            & 66.6 & 80.6 & 75.1 & 67.0 & 49.8 & 42.5 & 79.1 & 69.0 & 86.7 & 42.6 & 63.8 & 53.3 & 83.6 & 74.9 & 77.4 & 76.7 & 34.8 & 69.4 & 61.5 & 78.0 & 65.4 \\
\bottomrule[0.1 em]
\end{tabular}
\vspace{-1.0mm}
\caption{Object detection average precision ($\%$) on the PASCAL VOC 2012 \textit{val} set.}
\label{tab:perf_obj_det}
\end{table}

\newcommand{\twoline}[2]{ \parbox{\textwidth / 25}{ \centering #1\\ \centering #2 } }
\begin{table} [!bp]
\centering
\rowcolors{2}{}{gray!35}
\addtolength{\tabcolsep}{-5.7pt}
\scriptsize
\begin{tabular}{ c | c  c c c c c c c c c c c c c c c c c c c c }
\toprule[0.2 em] %
  Method &  mAP &  \twoline{areo}{body} & \twoline{areo}{stern} & \twoline{aero}{lwin} &  \twoline{aero}{rwin} &  \twoline{bike}{head} &  \twoline{bike}{body} &  \twoline{bike}{fwhe} & \twoline{bike}{bwhe}  &  \twoline{bird}{head} & \twoline{bird}{torso} & \twoline{bird}{neck} & \twoline{bird}{lwin} & \twoline{bird}{rwin} & \twoline{bird}{legs} & \twoline{bird}{tail} & \twoline{boat}{head} & \twoline{boat}{body} & \twoline{boat}{rear} & \twoline{boat}{sail} & \twoline{bottl}{cap}  \\
  \textit{op rcnn}& 22.0 & 42.5 & 11.3 & 2.8 & 2.9 & 3.3  & 36.4 & 22.9 & 20.1 & 9.8  & 45.8 & 2.6 & 1.4 & 2.6 & 3.4 & 0.9 & 0.7 & 11.3 & 0.1 & 33.0 & 0.9 \\
  DeePM           & 24.9 & 52.7 & 40.3 & 4.1 & 5.7 & 13.6 & 38.1 & 35.6 & 32.6 & 35.7 & 44.5 & 2.3 & 3.1 & 5.1 & 8.6 & 4.8 & 5.2 & 18.1 & 1.0 & 25.0 & 1.5 \\
 \midrule
 \midrule
  Method & \twoline{bottl}{body} & \twoline{bus}{front} & \twoline{bus}{lsid} & \twoline{bus}{rsid} & \twoline{bus}{back} & \twoline{car}{fron} & \twoline{car}{lsid} &  \twoline{car}{rsid} & \twoline{car}{back} & \twoline{cat}{head} & \twoline{cat}{torso} & \twoline{cat}{fleg} & \twoline{cat}{bleg} & \twoline{cat}{tail} & \twoline{chair}{back} & \twoline{chair}{cushi} & \twoline{chair}{lside} & \twoline{chair}{rside} & \twoline{chair}{botto} & \twoline{cow}{head} & \twoline{cow}{torso} \\
  \textit{op rcnn}  & 39.0 & 65.9 & 53.8 & 60.8 & 2.0  & 37.2 & 30.1 & 28.0 & 16.4 & 55.7 & 54.4 & 8.0 & 1.2 & 0.5 & 17.3 & 9.1  & 0.7 & 2.2 & 27.3 & 16.2 & 39.8  \\
  DeePM             & 33.1 & 69.5 & 39.2 & 30.1 & 10.6 & 37.1 & 19.7 & 22.5 & 20.8 & 82.1 & 52.7 & 7.3 & 1.7 & 5.8 & 26.1 & 12.2 & 1.5 & 1.5 & 23.4 & 40.3 & 41.0 \\
 \midrule
 \midrule
  Method & \twoline{cow}{fleg} & \twoline{cow}{bleg} & \twoline{table}{top} & \twoline{table}{botto} & \twoline{dog}{head} & \twoline{dog}{torso}  & \twoline{dog}{fleg} & \twoline{dog}{bleg} & \twoline{dog}{tail} & \twoline{horse}{head} & \twoline{horse}{torso} & \twoline{horse}{neck} & \twoline{horse}{fleg} & \twoline{horse}{bleg} & \twoline{horse}{tail} & \twoline{mbik}{head} & \twoline{mbik}{body} & \twoline{mbik}{fwhe} & \twoline{mbik}{bwhe} & \twoline{mbik}{rear}   & \twoline{perso}{head} \\
  \textit{op rcnn} & 13.7 & 3.9 & 42.9 & 5.7 & 56.5 & 44.6 & 14.6 & 1.5 & 0.2 & 13.8 & 52.9 & 15.6 & 20.8 & 15.2 & 1.3  & 12.9 & 52.2 & 16.5 & 11.9 & 4.9  & 51.8 \\
  DeePM            & 8.4  & 4.1 & 43.7 & 2.3 & 75.7 & 47.5 & 19.9 & 4.4 & 7.1 & 44.8 & 51.5 & 17.7 & 17.8 & 14.8 & 11.3 & 26.4 & 42.4 & 30.0 & 15.4 & 11.8 & 51.3 \\
 \midrule
 \midrule
  Method &  \twoline{perso}{torso} & \twoline{perso}{lulim} & \twoline{perso}{rulim} & \twoline{perso}{llimb} & \twoline{plant}{pot} & \twoline{plant}{plant} & \twoline{shee}{head} & \twoline{shee}{torso} & \twoline{shee}{flegs} & \twoline{shee}{blegs} & \twoline{sofa}{back} & \twoline{sofa}{cushi} & \twoline{sofa}{front} & \twoline{sofa}{lside} & \twoline{sofa}{rside} & \twoline{train}{head} & \twoline{train}{front} & \twoline{train}{lside} & \twoline{train}{rside} & \twoline{train}{coach} & \twoline{tv}{scree} \\
  \textit{op rcnn} & 59.7 & 12.5 & 12.4 & 39.2 & 13.7 & 22.6 & 10.5 & 60.9 & 3.4 & 1.1 & 21.1 & 12.9 & 8.3  & 3.0 & 1.1 & 68.6 & 51.1 & 37.5 & 38.0 & 7.4  & 64.2 \\
  DeePM            & 46.0 & 11.4 & 10.8 & 30.7 & 22.1 & 20.5 & 29.8 & 64.0 & 1.0 & 0.0 & 28.4 & 18.3 & 12.4 & 6.8 & 9.1 & 66.5 & 56.7 & 19.8 & 26.4 & 16.9 & 64.7 \\
\bottomrule[0.1 em]
\end{tabular}
\vspace{-1.0mm}
\caption{Part detection average precision ($\%$) on the PASCAL VOC 2012 \textit{val} set. Please refer to Table \ref{tab:leeds} for the full name of each object semantic part.}
\label{tab:perf_part_det}
\end{table}

\begin{table} [!bp]
\centering
\rowcolors{2}{}{gray!35}
\addtolength{\tabcolsep}{-4.0pt}
\scriptsize
\begin{tabular}{ c | c | c c c c c c c c c c c c c c c c c c c c }
\toprule[0.2 em]
 Method &  mAP          & areo &  bike &  bird &  boat &  bottle &  bus &  car &  cat   & chair &  cow   & table &  dog &  horse & mbike &  person &  plant &  sheep &  sofa &  train &  tv \\
\textit{fast}    & 65.0 & 80.5 & 75.2  & 67.7  & 46.5    & 36.7  & 72.7 & 68.1 & 87.3   & 39.1  & 70.9   & 49.9  & 85.3 & 77.0   & 75.8  & 69.1    & 34.7   & 65.0   & 62.3  & 75.4   & 60.1 \\
\textit{faster}  & 66.8 & 80.1 & 76.0  & 70.0  & 49.7    & 44.9  & 73.4 & 72.9 & 87.3   & 40.0  & 73.5   & 50.0  & 86.8 & 79.8   & 79.3  & 75.7    & 34.1   & 69.2   & 57.8  & 77.2   & 59.0 \\
\midrule
\textit{op rcnn} & 67.4 & 82.2 & 76.6  & 70.9  & 49.1    & 44.8  & 73.3 & 72.2 & 88.6   & 42.8  & 72.9   & 49.8  & 86.9 & 81.1   & 78.5  & 78.1    & 35.9   & 70.8   & 59.6  & 76.6   & 57.3 \\
DeePM            & 67.7 & 82.6 & 76.2  & 70.3  & 49.9    & 44.7  & 72.9 & 72.0 & 88.7   & 43.8  & 73.3   & 51.5  & 87.0 & 80.6   & 78.1  & 76.2    & 36.3   & 71.0   & 61.7  & 76.2   & 61.6 \\
\bottomrule[0.1 em]
\end{tabular}
\vspace{-1.0mm}
\caption{Object detection average precision ($\%$) on the PASCAL VOC 2012 \textit{test} set.}
\label{tab:perf_obj_det_voc12test}
\end{table}

\newcommand{\fshort}[2]{#1 (\textit{#2})}
\begin{table*}[!b]
\centering
\rowcolors{2}{}{gray!35}
\addtolength{\tabcolsep}{-3.0pt}
\scriptsize
\begin{tabular}{ c | c | c | c | c | c | c }
\toprule[0.2 em] %
areo &  bike &  bird &  boat &  bottle &  bus &  car  \\
\fshort{body}{body}  & \fshort{head}{head} & \fshort{head}{head} & \fshort{head}{head} & \fshort{cap}{cap} & \fshort{front}{front} & \fshort{front}{fron} \\
\fshort{stern}{stern} & \fshort{body}{body} & \fshort{torso}{torso} & \fshort{body}{body} & \fshort{body}{body} & \fshort{left side}{lsid} & \fshort{left side}{lsid} \\
\fshort{left wing}{lwin} & \fshort{front wheel}{fwhe} & \fshort{neck}{neck} & \fshort{rear}{rear} & & \fshort{right side}{rsid} & \fshort{right side}{rsid} \\
\fshort{right wing}{rwin} & \fshort{back wheel}{bwhe} & \fshort{left wing}{lwin} & \fshort{sail}{sail} & & \fshort{back}{back} & \fshort{back}{back} \\
                   							& & \fshort{right wing}{rwin}  \\
		    							& & \fshort{legs}{legs} \\
		    							& & \fshort{tail}{tail} \\
 \midrule
 \midrule
cat &  chair &  cow & table & dog &  horse &  mbike \\
\fshort{head}{head}  & \fshort{back}{back} & \fshort{head}{head} & \fshort{top}{top} & \fshort{head}{head} & \fshort{head}{head} & \fshort{head}{head} \\
\fshort{torso}{torso} & \fshort{cushion}{cushi} & \fshort{torso}{torso} & \fshort{bottom}{botto} & \fshort{torso}{torso} & \fshort{torso}{torso} &\fshort{body}{body} \\
\fshort{front legs}{fleg} & \fshort{left side}{lside} & \fshort{front legs}{fleg} & & \fshort{front legs}{fleg} & \fshort{front legs}{fleg} & \fshort{front wheels}{fwhee} \\
\fshort{back legs}{bleg} & \fshort{right side}{rside} & \fshort{back legs}{bleg} & & \fshort{back legs}{bleg} & \fshort{back legs}{bleg} & \fshort{back wheels}{bwhee}  \\
\fshort{tail}{tail} & \fshort{bottom}{bottom}  & & & \fshort{tail}{tail} & \fshort{tail}{tail} & \fshort{rear}{rear} \\
 \midrule
 \midrule
person &  plant &  sheep &  sofa &  train &  tv \\
\fshort{head}{head} & \fshort{pot}{pot} & \fshort{head}{head}  & \fshort{back}{back} & \fshort{head}{head} & \fshort{screen}{scree} \\
\fshort{torso}{torso} &  \fshort{plant}{plant} & \fshort{torso}{torso} & \fshort{cushion}{cushi} & \fshort{front}{front} \\
\fshort{left upper limb}{lulim} & \fshort{front legs}{flegs} & \fshort{front}{front} & & \fshort{left side}{lside} \\
\fshort{right upper limg}{rulim} & \fshort{back legs}{blegs} & \fshort{left side}{lside} &  & \fshort{right side}{rside} \\
\fshort{lower limb}{llimb} & & \fshort{right side}{rside} &  & \fshort{coach}{coach} \\
\bottomrule[0.1 em]
\end{tabular}
\vspace{-1.0mm}
\caption{The full list of our part definition for the 20 PASCAL VOC object classes. The abbreviation for each part is in the parentheses. }
\label{tab:leeds}
\end{table*}

\begin{figure*}[!bp]
\begin{center}
	\includegraphics[width=\linewidth]{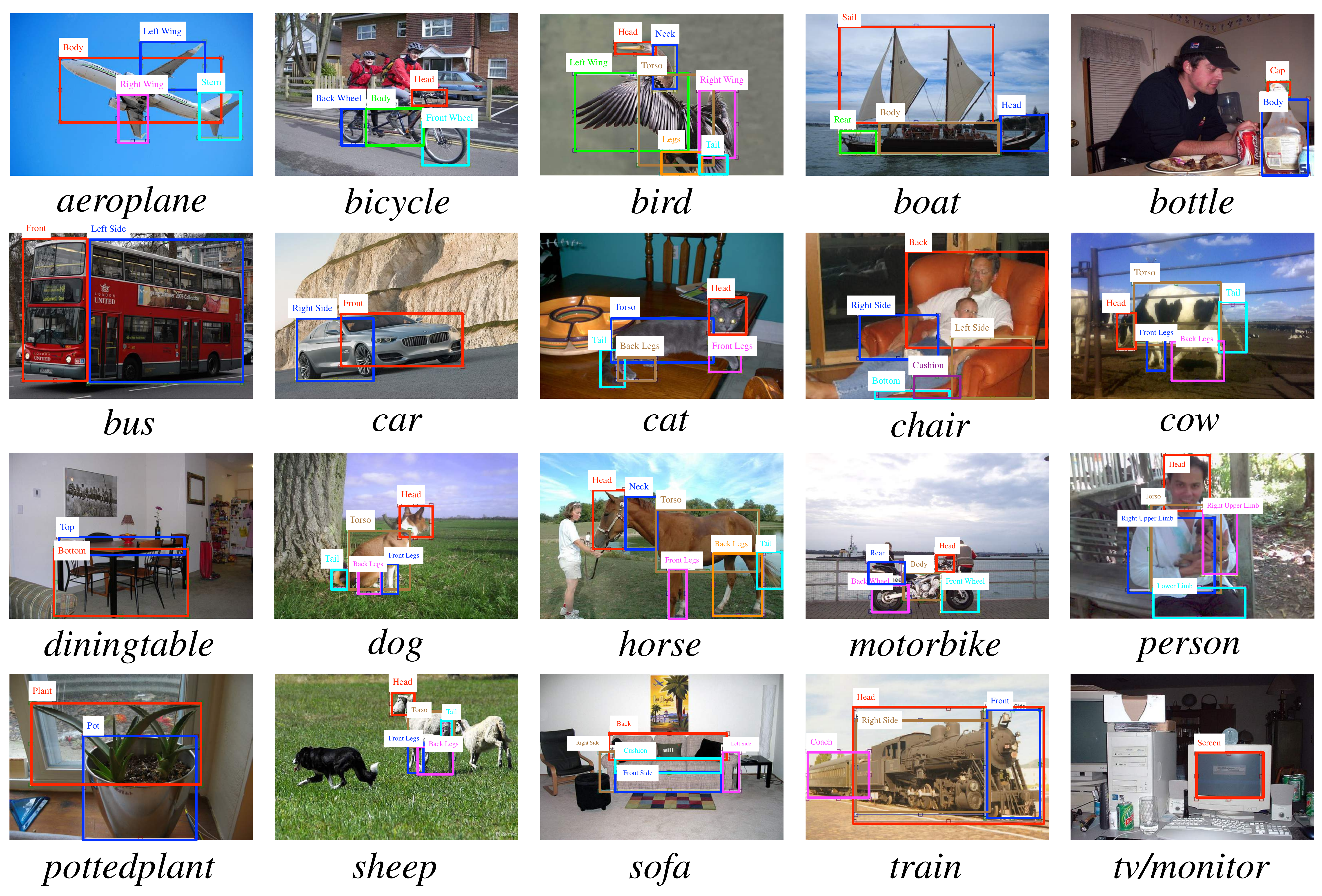}
\end{center}
   \caption{Illustration on our annotations of semantic object parts for the PASCAL VOC 2012 dataset. In this paper, we define 83 semantic part classes for all the 20 PASCAL VOC object classes. For clarity, we only visualize the part annotations for one object instance in each class. There may be multiple instances of the same object class in one image (e.g., the pictures of sheep and cow) and we have actually labelled the parts for each object instance (best viewed in color).}
\label{fig:anno}
\end{figure*}

\begin{figure*}[!]
\begin{center}
	\includegraphics[width=\linewidth]{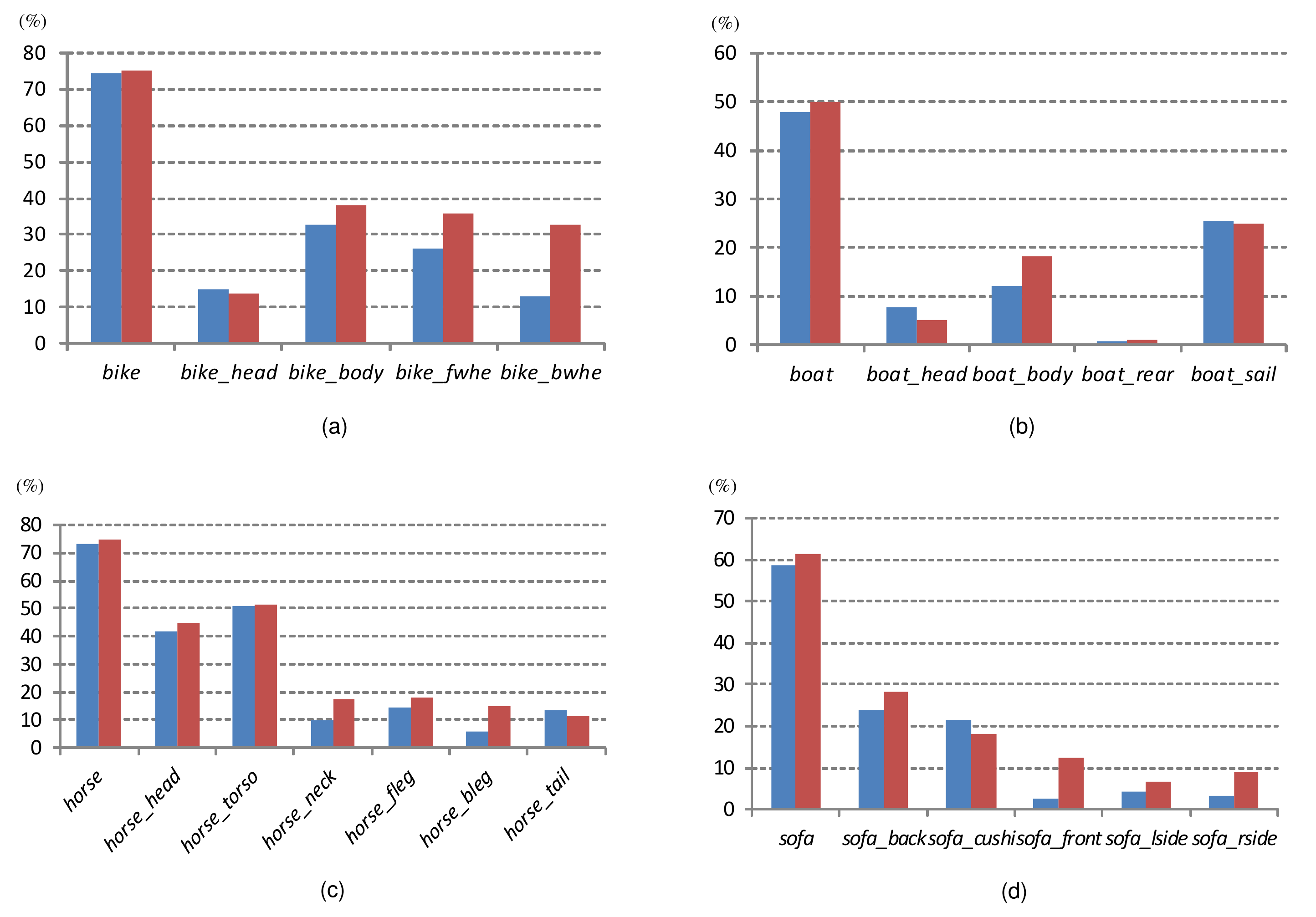}
\end{center}
   \caption{Performance (AP) comparison between DeePM (5 types used for the object node, 3 types used for each part node) and the `DPM-v1'-like single type baseline model. The blue and red color bars correspond to the baseline model and DeePM, respectively. (a) object class \textit{bicycle} and its parts; (b) object class \textit{boat} and its parts; (c) object class \textit{horse} and its parts; (d) object class \textit{sofa} and its parts. Please refer to table~\ref{tab:leeds} for the full names of object parts (best viewed in color).}
\label{fig:compr_single_type_model}
\end{figure*}

\begin{figure*}[!]
\begin{center}
	\includegraphics[width=\linewidth]{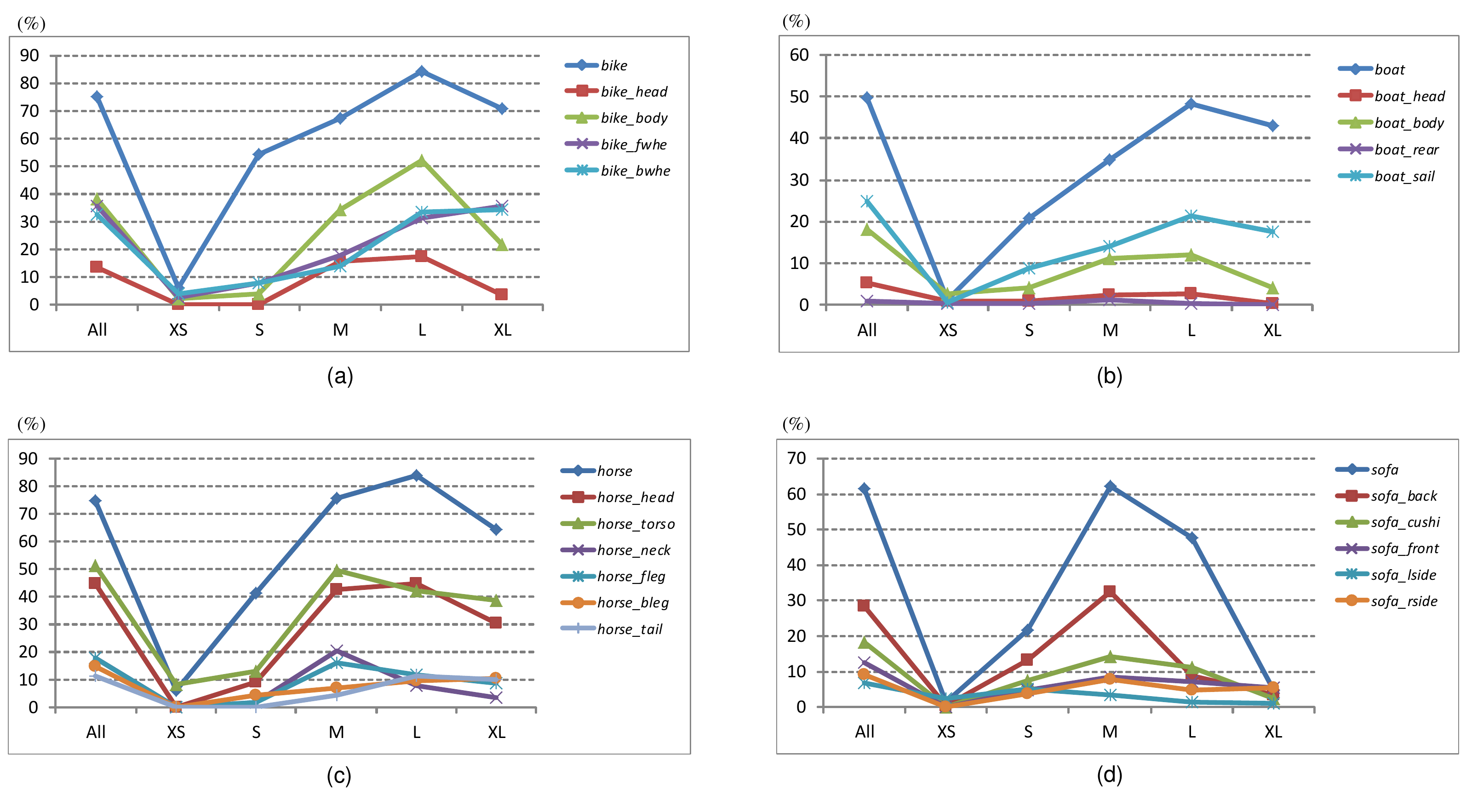}
\end{center}
   \caption{Performance (AP) w.r.t. different object/part size levels. `All': all examples; `XS': extremely small size; `S': small size; `M': medium size; `L': large size; `XL': extremely large size. (a) object class \textit{bicycle} and its parts; (b) object class \textit{boat} and its parts; (c) object class \textit{horse} and its parts; (d) object class \textit{sofa} and its parts. Please refer to table~\ref{tab:leeds} for the full names of object parts (best viewed in color).}
\label{fig:compr_perf_wrt_size}
\end{figure*}

\begin{figure*}[!]
\begin{center}
	\includegraphics[width=\linewidth]{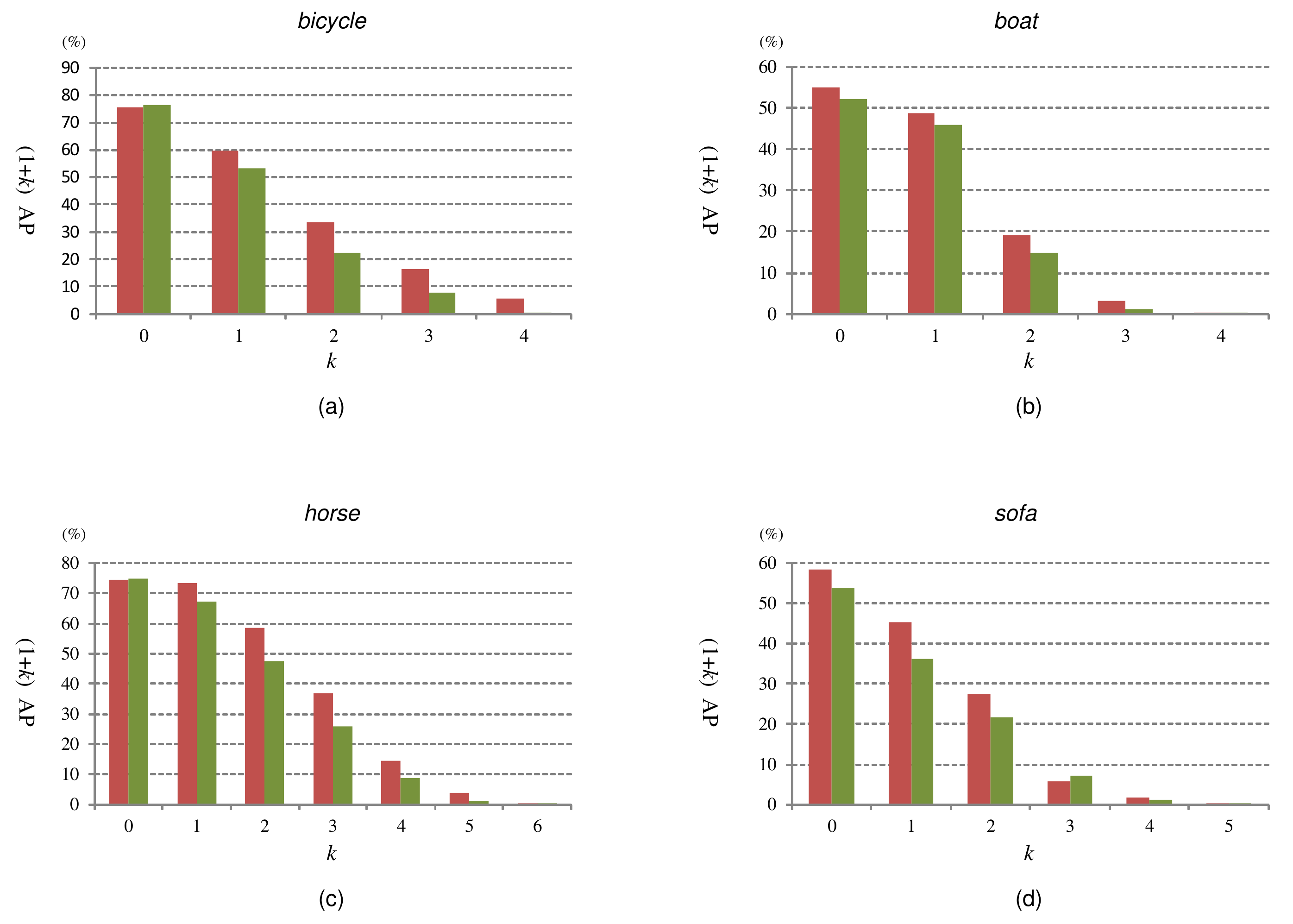}
\end{center}
   \caption{The ``$(1+k)$" average precision ($\%$) comparison for DeePM and OP R-CNN. The red and green color bars correspond to DeePM and OP R-CNN, respectively. (a) \textit{bicycle}; (b) \textit{boat}; (c) \textit{horse}; (d) \textit{sofa} (best viewed in color).}
\label{fig:compr_one_plus_k_aps}
\end{figure*}

\begin{figure*}[!]
\begin{center}
	\includegraphics[width=\linewidth]{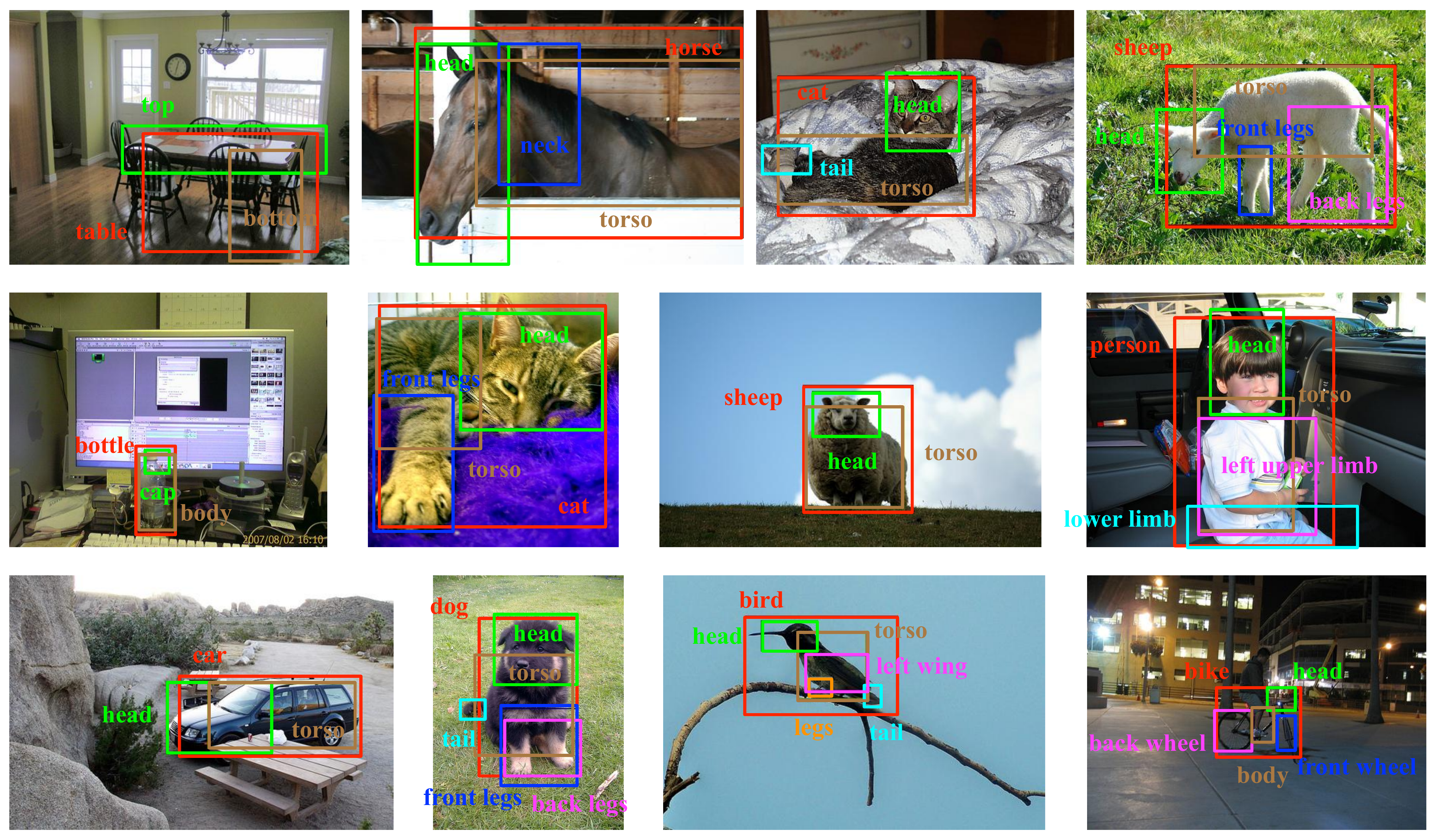}
\end{center}
   \caption{Visualization on some examples of the detected object-part configurations by DeePM (best viewed in color).}
\label{fig:vis_det}
\end{figure*}

\begin{figure*}[!]
\begin{center}
	\includegraphics[width=\linewidth]{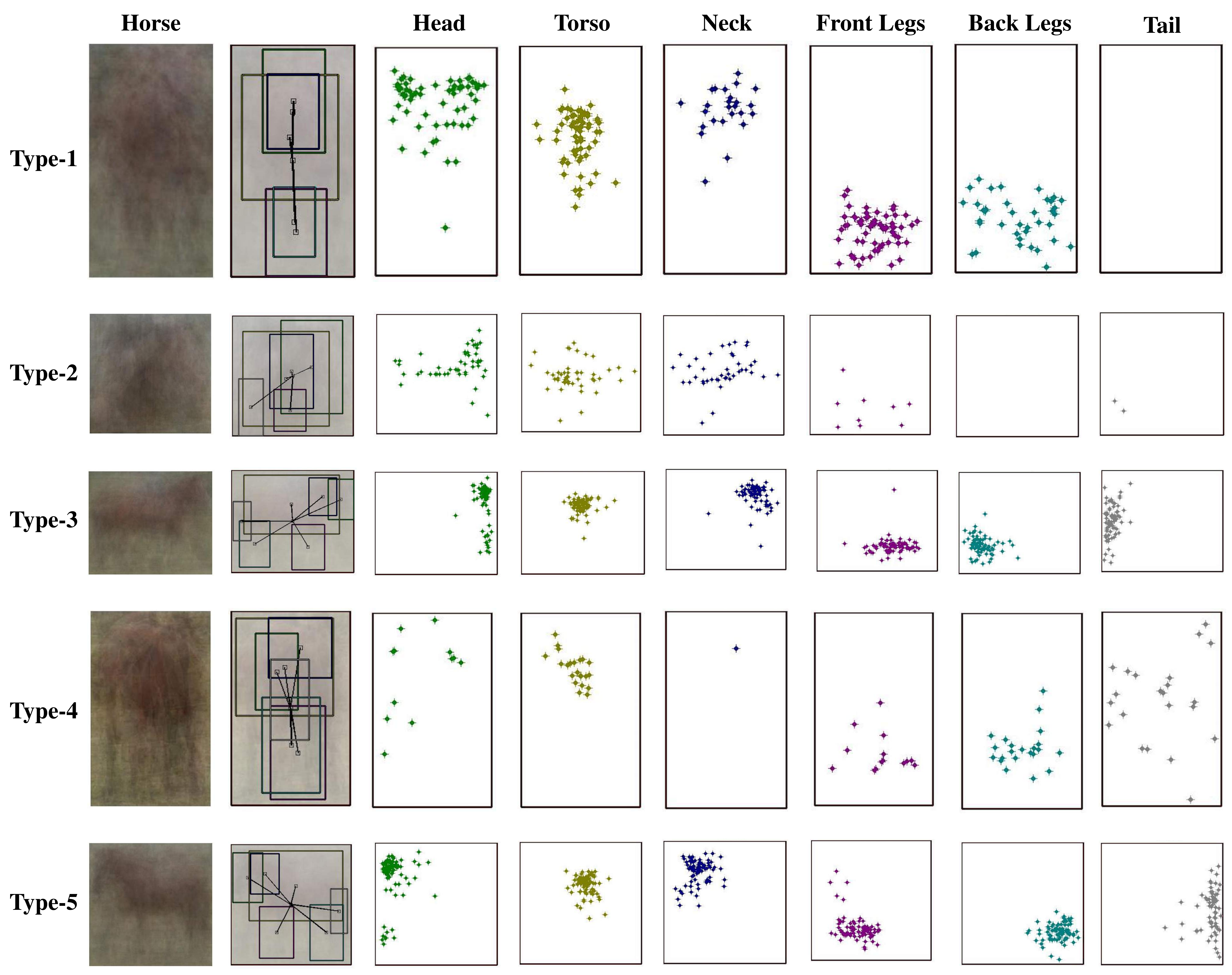}
\end{center}
   \caption{Visualization on the learned object types for \textit{horse} class. The first column is the average image over the examples of different types. The second column shows the anchor part bounding boxes (i.e., the mean bounding boxes over corresponding part instances) within the object bounding box. The rest columns visualize the normalized center locations of part instances w.r.t. the object bounding box for all the part classes, respectively (best viewed in color).}
\label{fig:vis_type_clusters_horse}
\end{figure*}

\begin{figure*}
\begin{center}
	\includegraphics[width=\linewidth]{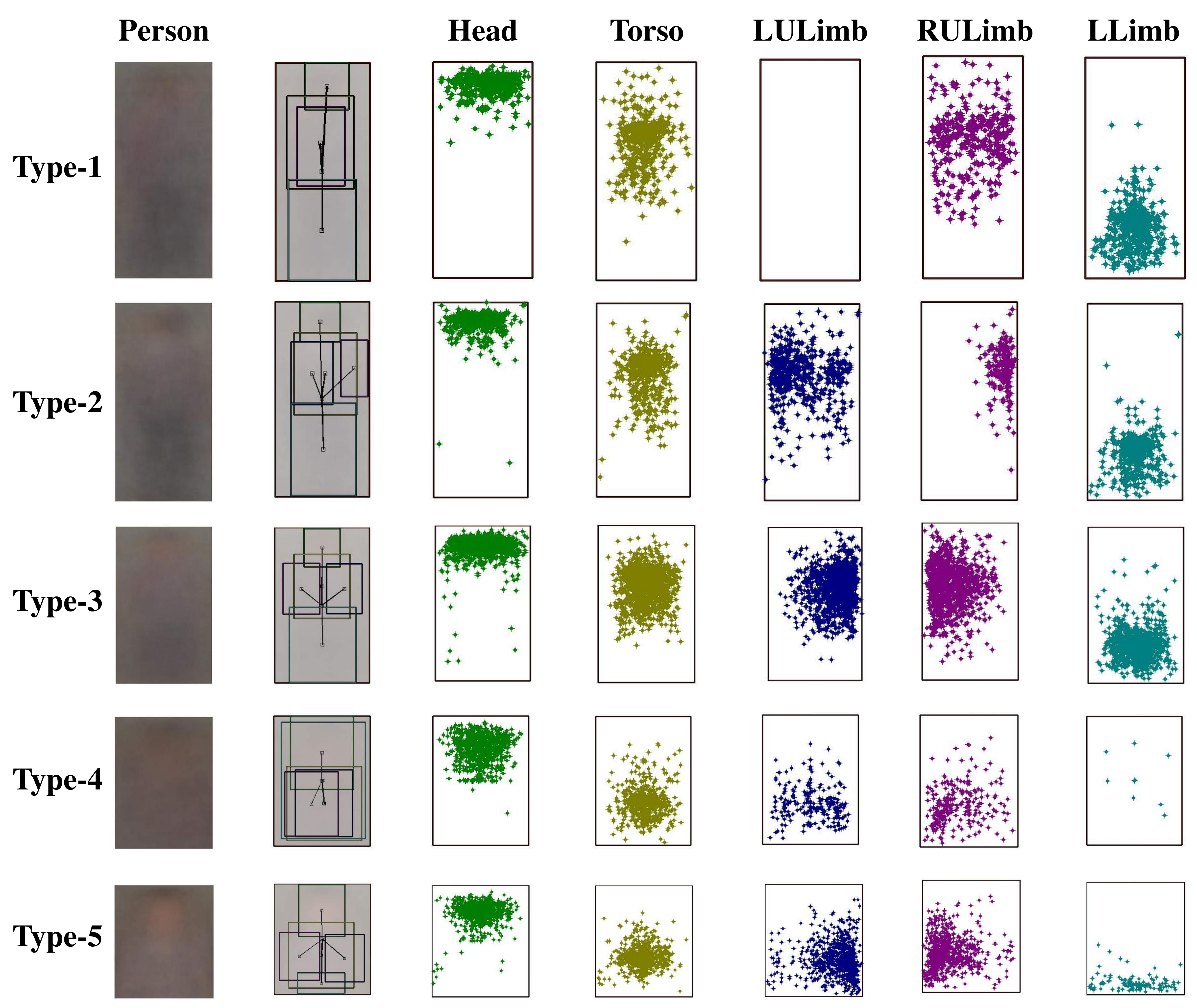}
\end{center}
   \caption{Visualization on the learned object types for \textit{person} class. The first column is the average image over the examples of different types. The second column shows the anchor part bounding boxes (i.e., the mean bounding boxes over corresponding part instances) within the object bounding box. The rest columns visualize the normalized center locations of part instances w.r.t. the object bounding box for all the part classes, respectively (best viewed in color).}
\label{fig:vis_type_clusters_person}
\end{figure*}

\end{document}